\documentclass[conference]{IEEEtran}
\IEEEoverridecommandlockouts
\usepackage{cite}
\usepackage{amsmath,amssymb,amsfonts}
\usepackage{algorithmic}
\usepackage{graphicx}
\usepackage{textcomp}
\usepackage{xcolor}
\usepackage{siunitx}
\usepackage{tcolorbox}
\usepackage{mathtools}

\DeclareMathOperator*{\argmax}{arg\,max}

\usepackage{forloop}
\newcounter{loopc}

\def\BibTeX{{\rm B\kern-.05em{\sc i\kern-.025em b}\kern-.08em
    T\kern-.1667em\lower.7ex\hbox{E}\kern-.125emX}}
\begin{document}

\title{
Active Label Refinement for Semantic Segmentation of Satellite Images
}

\author{\IEEEauthorblockN{Tuan Pham Minh}
\IEEEauthorblockA{\textit{Intelligent Embedded Systems} \\
\textit{University of Kassel}\\
Kassel, Germany \\
tuan.pham@uni-kassel.de}
\and
\IEEEauthorblockN{Jayan Wijesingha}
\IEEEauthorblockA{\textit{Grassland Science \& Renewable Plant Resources} \\
\textit{University of Kassel}\\
Witzenhausen, Germany \\
jayan.wijesingha@uni-kassel.de} 
\and
\IEEEauthorblockN{Daniel Kottke}
\IEEEauthorblockA{\textit{Intelligent Embedded Systems} \\
\textit{University of Kassel}\\
Kassel, Germany \\
daniel.kottke@uni-kassel.de}
\and
\IEEEauthorblockN{Marek Herde}
\IEEEauthorblockA{\textit{Intelligent Embedded Systems} \\
\textit{University of Kassel}\\
Kassel, Germany \\
marek.herde@uni-kassel.de}
\and
\IEEEauthorblockN{Denis Huseljic}
\IEEEauthorblockA{\textit{Intelligent Embedded Systems} \\
\textit{University of Kassel}\\
Kassel, Germany \\
dhuseljic@uni-kassel.de}
\and
\IEEEauthorblockN{Bernhard Sick}
\IEEEauthorblockA{\textit{Intelligent Embedded Systems} \\
\textit{University of Kassel}\\
Kassel, Germany \\
bsick@uni-kassel.de}
\and
\IEEEauthorblockN{Michael Wachendorf}
\IEEEauthorblockA{\textit{Grassland Science \& Renewable Plant Resources} \\
\textit{University of Kassel}\\
Witzenhausen, Germany \\
mwach@uni-kassel.de} 
\and
\IEEEauthorblockN{Thomas Esch}
\IEEEauthorblockA{\textit{German Remote Sensing Data Center} \\
\textit{German Aerospace Center}\\
Oberpfaffenhofen, Germany \\
thomas.esch@dlr.de}
}

\maketitle

\begin{abstract}
Remote sensing through semantic segmentation of satellite images contributes to the understanding and utilisation of the earth's surface. For this purpose, semantic segmentation networks are typically trained on large sets of labelled satellite images. However, obtaining expert labels for these images is costly. Therefore, we propose to rely on a low-cost approach, e.g. crowdsourcing or pretrained networks, to label the images in the first step. Since these initial labels are partially erroneous, we use active learning strategies to cost-efficiently refine the labels in the second step. We evaluate the active learning strategies using satellite images of Bengaluru in India, labelled with land cover and land use labels. Our experimental results suggest that an active label refinement to improve the semantic segmentation network's performance is beneficial.
\end{abstract}


\section{Introduction}

\begin{figure}[!t]
    \includegraphics[width=\columnwidth]{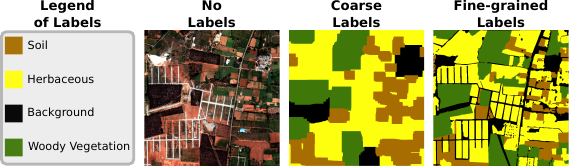}
    \caption{Illustration of semantic segmentation for satellite images with four classes: The three images show the same satellite image at three different labelling stages, i.e. without any labels (unlabelled), with coarse labels (e.g. obtained via crowdsourcing or from a pretrained semantic segmentation network), and with fine-grained labels (e.g. obtained from an expert).}
    \label{fig:label-overview}
\end{figure}

\textbf{Semantic segmentation}, a pivotal task involving pixel-level image categorisation, holds immense significance within remote sensing (e.g. land-cover mapping)~\cite{guo2018review, wurm2019semantic}. This task is inherently intricate, and despite the considerable efficacy of semantic segmentation networks, their susceptibility to shortcomings persists. Even though acceptable results are shown with benchmark datasets, the performance of semantic segmentation networks on real-life datasets still needs to improve~\cite{jiang2022intelligent}. Compared to natural images, remote sensing images, e.g. satellite images, exhibit inherent challenges arising from high spectral heterogeneity and complex scenarios, including image occlusions and artefacts~\cite{jiang2022intelligent}.
Furthermore, the scarcity of extensively labelled training data for specific applications compounds the complexity. Labelling images in semantic segmentation problems is an intricate and labour-intensive process, demanding expert labellers to meticulously demarcate object boundaries and areas of interest meticulously~\cite{neupane2021deep}. Fig.~\ref{fig:label-overview} gives a corresponding example of labelling a satellite image. Flaws or inconsistencies in these labels (also known as coarse or noisy labels~\cite{acuna2019devil}) can significantly impair the performance of semantic segmentation networks, often resulting in misclassifications and reduced generalisation capabilities~\cite{lenczner2022dial}. 

\textbf{Active learning}~\cite{settles2009active,herde2021survey} is a promising machine learning paradigm for handily addressing the difficulties of obtaining high-quality labelled data. In the context of semantic segmentation, an active learning strategy aims to maximise the semantic segmentation network's performance while reducing labelling costs. Therefor, such a strategy selects images whose labellings are expected to be most beneficial for training. 
Recently, Desai et al.~\cite{desai2022active} proposed an active learning strategy to select a representative collection of images for labelling, which improved the semantic segmentation network's performance by sizeable margins compared to a random selection. Similarly, Lenczner et al.~\cite{lenczner2022dial} introduced the deep interactive and active learning framework for semantic segmentation, which guides users to refine the labels of the most relevant image areas.

Our scenario is similar to that of Lenczner et al.~\cite{lenczner2022dial} and is sketched in Fig.~\ref{fig:al-cycle}. Unlike classical active learning, we start with an already labelled dataset and focus on satellite images. The initial labels are coarse and contain labelling errors, possibly due to human error or labelling with a semantic segmentation network trained on images from different spatial and/or temporal domains. Using these initial labels, we can train a semantic segmentation network. An active learning strategy selects the best areas among randomly generated equally-sized areas. The selected areas are then relabelled by an expert and used as training data for the next active learning cycle. Accordingly, we aim to answer the following \textbf{research question} in this article:
\begin{tcolorbox}
    Can active learning identify coarsely labelled areas of satellite images where label refinement maximises a semantic segmentation network's performance? 
\end{tcolorbox}

\section{Data}
\label{sec:data}
\subsection{Data Showcase}
The dataset used in this article is a cloud-free WorldView-3 \textit{satellite visible and near-infrared} (VNIR) image captured in 2018. The satellite image covers an urban-rural transect of the northern part of Bengaluru, India. The image has eight spectral bands with a spatial resolution of \SI{1.24}{\metre}. The raw digital numbers were converted to at-sensor radiance (\unit{\watt\per\square\meter\per\micro\meter}) and then to at-sensor reflectance (in the interval $[0, 1]$). However, for this article, only five bands were selected: blue, green, red, red-edge and near-infrared. From the whole image, only twelve \SI{1}{\kilo\metre\squared} regions ($833 \times 833$ pixels) were extracted as smaller satellite images along the urban-rural gradient of Bengaluru.

Each \SI{1}{\kilo\metre\squared} satellite image is manually digitised as vector polygons by the expert into four classes: background, soil, herbaceous, and woody vegetation (cf. fine-grained labels in~Fig.~\ref{fig:label-overview}). The digitised polygons are converted into rasters with the same pixel size as the satellite image. The soil class explains the barren lands and fallow lands. The orchards (e.g. pomegranate, mango) and plantations (e.g. eucalyptus) are considered woody vegetation. The agricultural crop fields (e.g. millet, maize) are labelled herbaceous vegetation.

\subsection{Challenges}

Given the spatial resolution of \SI{1.24}{\metre} in the satellite images, the challenge in performing semantic segmentation lies in the presence of multiple small objects within the target classes. This complexity was particularly pronounced when attempting to segment these classes accurately. Simultaneously, the task of assigning accurate labels to these small objects introduced additional difficulties. For example, the woody vegetation class encompassed areas like plantations, which comprised tree crowns and patches of soil or grass between rows of trees. The inter-class similarity was another difficulty. The uncultivated agricultural lands were usually labelled as the soil class. However, some unpaved roads (considered background class) also showed similar soil colours and textures. Analogously, many further similarities exist between the other three classes and the background class. Moreover, the data exhibit intra-class dissimilarities, most evident in the herbaceous and woody vegetation classes. For example, the herbaceous class encompassed croplands housing various field and vegetable crops, thereby introducing variations within the class.

\section{Active Label Refinement}
\subsection{Scenario}
\label{sec:method:scenario}
Labelling images for semantic segmentation is a complex task in itself. The high amount of data required to solve remote sensing problems results in a time-consuming process that might require experts, depending on the task. Thus, we want to explore using active learning to improve coarse but cheaply generated labelled data. These coarse labels might be obtained via crowdsourcing or a semantic segmentation network pretrained on different data. 
Thus, contrary to classical active learning, where we typically start with an empty or very small pool of labelled data, our scenario assumes that we have a fully labelled dataset. Our task in this scenario is to find areas where the initially coarse labels might be wrong so an expert can refine these areas to maximize the semantic segmentation network's performance.

\begin{figure}[!b]
    \centering
    \includegraphics[width=\columnwidth]{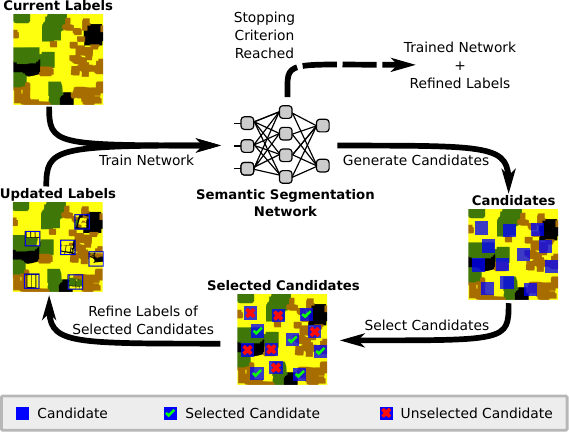}
    \caption{Illustration of an active learning cycle: Starting from the current (partially coarsely) labelled satellite images, a semantic segmentation network is trained to intelligently select candidate areas for label refinement. Such an active learning cycle is repeated until a stopping criterion is met (e.g. exhausting the label budget).}
    \label{fig:al-cycle}
\end{figure}

\subsection{Semantic Segmentation Network}
The semantic segmentation network we use in this article is based on a ResNet-50~\cite{he2016deep} pretrained on ImageNet~\cite{deng2009imagenet}. As our dataset consists of more than three bands, i.e. red, green and blue, we employ two pretrained ResNet-50 models as a backbone. The first backbone network is fed the red, green and blue bands, while the second one is fed red-edge, near-infrared and the average from the red, green and blue bands. Each backbone network provides four outputs, which we aggregate through a convolution layer using the \textit{rectified linear unit} (ReLU)~\cite{glorot2011deep} as activation function and batch normalisation~\cite{ioffe2015batch}. As a result, the outputted images have the same size as the input images with four channels, i.e. one for each class (cf. Fig.~\ref{fig:label-overview}). The semantic segmentation network is trained using $128 \times 128$ image chips randomly sampled from the twelve $833 \times 833$ satellite images. Furthermore, we use a weighted cross-entropy loss function and Adam~\cite{kingma2015adam} to optimise the semantic segmentation network's weights.

\subsection{Simulation of Coarse Labels}
To investigate our scenario, we need two different types of labels. For this article, we transform our available fine-grained labels to generate a new set of coarse labels, which we will use as initial labels. Our main goal for this transformation is to remove fine-grained labels but keep the original labels for large areas that are easily identified correctly. The basic idea behind such a transformation is to replace the original labels with labels where a single class is enlarged. By doing this repeatedly for all classes, small details are overridden by these operations. We implement the enlargement step via two-dimensional convolutions, for which we generate a randomly sized rectangular filter filled with only ones. The sizes for these filters range from $2 \times 2$ to $32 \times 32$ pixels. Using this filter, we can identify all pixels near other pixels classified as the class we are enlarging and relabel them.

\begin{figure}[!h]
    \centering
    \includegraphics[width=\columnwidth]{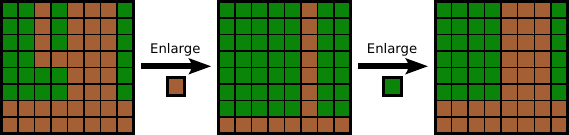}
    \caption{Simple example of our coarse label simulation: First, the herbaceous class is enlarged, followed by an enlargement of the soil class. Both enlargement steps are done with a $3 \times 3$ convolution filter.}
    \label{fig:label_augmentation}
\end{figure}

\subsection{Active Learning Strategies}
In each active learning cycle of Fig.~\ref{fig:al-cycle}, we randomly sample $N \in \mathbb{N}_{>0}$ candidates as satellite image areas, which we denote as $\mathcal{X} \coloneqq \{\boldsymbol{x}_1, \dots, \boldsymbol{x}_N\} \subset \mathbb{R}^{C \times W \times H}$ with $C, W, H \in \mathbb{N}_{>0}$ as the image dimensions, i.e. number of channels, width and height. Then, the active learning strategy selects a subset \mbox{$\mathcal{S} \coloneqq \{\boldsymbol{x}_{s_1}, \dots, \boldsymbol{x}_{s_K}\} \subset \mathcal{X}$}, \mbox{$\{s_1, \dots, s_K\} \subset [N] \coloneqq \{1, \dots, N\}$} of the \mbox{$K \in \mathbb{N}_{>0}$} most beneficial image areas. In our case, this selection is based on a utility function $U: \mathcal{X} \rightarrow \mathbb{R}$ identifying the image areas with the highest utility scores:
\begin{equation}
    \mathcal{S} \coloneqq \argmax_{\mathcal{S}^\prime \subset \mathcal{X}, |\mathcal{S}^\prime|=K}\left(\sum_{\boldsymbol{x} \in \mathcal{S}^\prime} U(\boldsymbol{x})\right).
\end{equation}
The pixel-level labels of the selected candidates are then refined by an expert as the basis for retraining the semantic segmentation network. Such an active learning cycle is repeated until a stopping criterion is met, e.g. reaching the maximum label budget.

Since the candidates $\mathcal{X}$ are sampled randomly, their respective number of pixels whose labels have been already refined might differ. Thus, we propose \textit{coverage sampling} (CS), an active learning strategy that maximises the number of pixels whose labels have not been refined yet.
For this purpose, we introduce the acquisition masks \mbox{$\mathcal{A} \coloneqq \{\mathbf{A}_1, \dots, \mathbf{A}_N\} \subset \{0, 1\}^{W \times H}$}. A pixel with the coordinates $(i, j) \in [W] \times [H]$ in a candidate area $\boldsymbol{x}_n \in \mathcal{X}$ that has been part of a label refinement is indicated by $A_n[i,j] = 0$ and by $A_n[i,j] = 1$, otherwise. Thus, CS calculates the utility score for a given candidate through:
\begin{equation}
    U_{\text{CS}}(\boldsymbol{x}_n) \coloneqq \sum_{i=1}^{W}\sum_{j=1}^H A_n[i,j].
\end{equation}

As another active learning strategy, we adapt \textit{uncertainty sampling} (US)~\cite{lewis1994heterogenous} to incorporate the coverage. Therefor, we compute the entropies \mbox{$\mathcal{H} \coloneqq \{\mathbf{H}_1, \dots, \mathbf{H}_N\}\subset \mathbb{R}_{\geq 0}^{W \times H}$} as uncertainty measurements, where $H_n[i,j]$ denotes the entropy of the segmentation network's class-membership predictions for pixel $(i, j) \in [W] \times [H]$ of candidate $\boldsymbol{x}_n \in \mathcal{X}$. The utility is computed by summing the entropies of not yet refined pixels:
\begin{equation}
    U_{\mathrm{US}}(\boldsymbol{x}_n) \coloneqq \sum_{i=1}^{W}\sum_{j=1}^H A_n[i,j] \cdot H_n[i,j].
\end{equation}

As a baseline, we also employ \textit{random sampling} (RS), which randomly selects candidates for label refinement.

\section{Experimental Evaluation}
\subsection{Experiments}
Given the twelve satellite images (cf. Section~\ref{sec:data}), we use a $20$ times repeated leave-one-out cross-validation to assess the active learning strategies' performances. The image chips used for training the semantic segmentation network are randomly sampled from the eleven training satellite images at each of the $15$ training epochs. Further, we utilise randomised image augmentations such as rotation, flipping and blurring. The semantic segmentation network is trained with the currently available labels, which are updated in each active learning cycle by incorporating the refined labels. After the training, the active learning strategy selects $K=16$ out of $N=128$ randomly sampled image areas.

\subsection{Evaluation}
\begin{figure}[!htb]
    \centering
    \includegraphics[width=\linewidth]{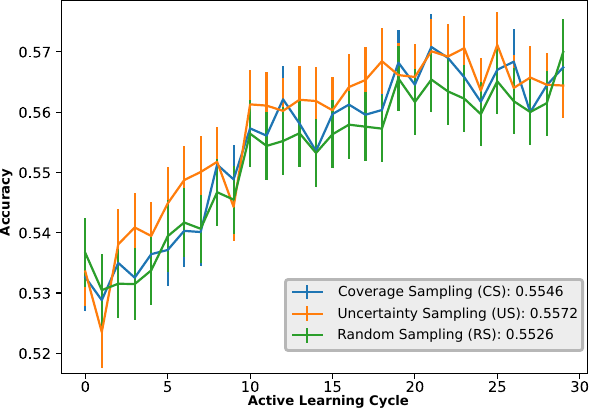}
    \caption{The curves show the mean accuracy and standard error per active learning cycle across all $20$ cross-validations for each tested strategy. The mean accuracy across the active learning cycles is given in the legend.}
    \label{fig:results:accuracy}
\end{figure}

Fig.~\ref{fig:results:accuracy} shows the results of our experiments. We can see an overall increase in performance after label refinement. US performs the best, while CS performs slightly better than RS.
The results show that CS behaves similarly to RS in the first cycles but performs better in the later cycles. The experiments start with solely coarse labels, i.e. without an expert's label refinements. Thus, the chance to present an image area to the expert with already labelled pixels is low in the early cycles but increases over time. Furthermore, we can observe that US improves upon CS even in the early cycles. Contrary to classical active learning, the training data does not increase in size over time. Thus, we do not have the problem of training the semantic segmentation network with only a few training images. We assume this is particularly beneficial for US, which focuses on exploiting the current decision boundary. As the initial labelled training set is similar to the actual ground truth, exploration is needed much less due to the high amount of correctly labelled training data early on.

\begin{figure}[!t]
    \centering
    \includegraphics[width=\linewidth]{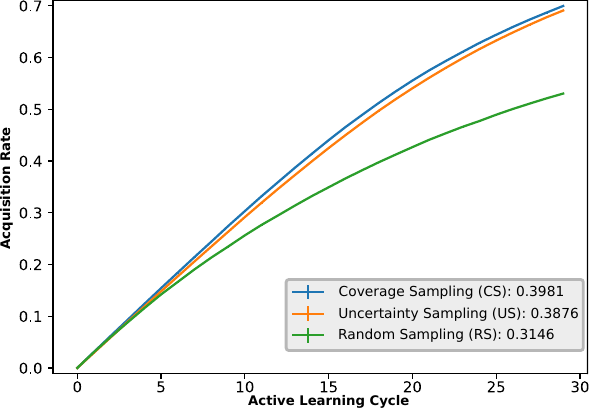}
    \caption{The curves show the mean accuracy and standard deviation per active learning cycle across all $20$ cross-validations for each tested strategy. The mean acquisition rate across the active learning cycles is given in the legend.}
    \label{fig:results:acquisition_ratios}
\end{figure}

Even though US performs better than CS, we can see in Fig.~\ref{fig:results:acquisition_ratios} that it queries slightly fewer new pixels. CS and US reach a plateau at cycle $20$, while RS performs similarly to the other strategies at cycles $29$ and $30$. These cycles point toward an approximate acquisition rate of \SI{50}{\percent}, i.e. the expert has refined half of the available pixels.
This indicates that US can identify image areas that need label refinement. We assume this happens as predictions for small objects misclassified in the initial labelling are ambiguous. Thus, they have higher uncertainty and are more likely to be selected.

\section{Conclusion}
In this article, we indicated that active learning can improve the semantic segmentation network by identifying areas where label refinement by an expert is beneficial. Future work may apply this scenario to benchmark data or use active learning strategies that select areas based on labelling errors by comparing predictions against initially coarse labels. Moreover, we need to account for the actual labelling costs, which are often correlated to the labelling time of an expert. Accordingly, the number of image areas selected for label refinement is not sufficient. Instead, we need to count the number of and estimate the complexity of the pixels selected for label refinement.

\bibliographystyle{ieeetran}
\bibliography{main}

\begin{thebibliography}{10}
\providecommand{\url}[1]{#1}
\csname url@samestyle\endcsname
\providecommand{\newblock}{\relax}
\providecommand{\bibinfo}[2]{#2}
\providecommand{\BIBentrySTDinterwordspacing}{\spaceskip=0pt\relax}
\providecommand{\BIBentryALTinterwordstretchfactor}{4}
\providecommand{\BIBentryALTinterwordspacing}{\spaceskip=\fontdimen2\font plus
\BIBentryALTinterwordstretchfactor\fontdimen3\font minus
  \fontdimen4\font\relax}
\providecommand{\BIBforeignlanguage}[2]{{%
\expandafter\ifx\csname l@#1\endcsname\relax
\typeout{** WARNING: IEEEtran.bst: No hyphenation pattern has been}%
\typeout{** loaded for the language `#1'. Using the pattern for}%
\typeout{** the default language instead.}%
\else
\language=\csname l@#1\endcsname
\fi
#2}}
\providecommand{\BIBdecl}{\relax}
\BIBdecl

\bibitem{guo2018review}
Y.~Guo, Y.~Liu, T.~Georgiou, and M.~S. Lew, ``{A review of semantic
  segmentation using deep neural networks},'' \emph{Int. J. Multimed. Inf.
  Retr.}, vol.~7, no.~2, pp. 87--93, 2018.

\bibitem{wurm2019semantic}
M.~Wurm, T.~Stark, X.~X. Zhu, M.~Weigand, and H.~Taubenb{\"{o}}ck, ``{Semantic
  segmentation of slums in satellite images using transfer learning on fully
  convolutional neural networks},'' \emph{ISPRS J. Photogramm. Remote Sens.},
  vol. 150, pp. 59--69, 2019.

\bibitem{jiang2022intelligent}
B.~Jiang, X.~An, S.~Xu, and Z.~Chen, ``{Intelligent Image Semantic
  Segmentation: A Review Through Deep Learning Techniques for Remote Sensing
  Image Analysis},'' \emph{J. Indian. Soc. Remote Sens.}, 2022.

\bibitem{neupane2021deep}
B.~Neupane, T.~Horanont, and J.~Aryal, ``{Deep Learning-Based Semantic
  Segmentation of Urban Features in Satellite Images: A Review and
  Meta-Analysis},'' \emph{Remote Sens.}, vol.~13, no.~4, p. 808, 2021.

\bibitem{acuna2019devil}
D.~Acuna, A.~Kar, and S.~Fidler, ``{Devil is in the Edges: Learning Semantic
  Boundaries from Noisy Annotations},'' in \emph{IEEE/CVF Conf. Comput. Vis.
  Pattern Recognit.}, Long Beach, CA, 2019, pp. 11\,075--11\,083.

\bibitem{lenczner2022dial}
G.~Lenczner, A.~Chan-Hon-Tong, B.~{Le Saux}, N.~Luminari, and G.~{Le
  Besnerais}, ``{DIAL: Deep Interactive and Active Learning for Semantic
  Segmentation in Remote Sensing},'' \emph{IEEE J. Sel. Top. Appl. Earth Obs.
  Remote Sens.}, vol.~15, pp. 3376--3389, 2022.

\bibitem{settles2009active}
B.~Settles, ``{Active Learning Literature Survey},'' University of Wisconsin --
  Madison, Computer Sciences Technical Report 1648, 2009.

\bibitem{herde2021survey}
M.~Herde, D.~Huseljic, B.~Sick, and A.~Calma, ``{A Survey on Cost Types,
  Interaction Schemes, and Annotator Performance Models in Selection Algorithms
  for Active Learning in Classification},'' \emph{IEEE Access}, vol.~9, pp.
  166\,970--166\,989, 2021.

\bibitem{desai2022active}
S.~Desai and D.~Ghose, ``{Active Learning for Improved Semi-Supervised Semantic
  Segmentation in Satellite Images},'' in \emph{IEEE/CVF Winter Conf. Appl.
  Comput. Vis.}, Waikoloa, HI, 2022, pp. 553--563.

\bibitem{he2016deep}
K.~He, X.~Zhang, S.~Ren, and J.~Sun, ``{Deep Residual Learning for Image
  Recognition},'' in \emph{Conf. Comput. Vis. Pattern Recognit.}, Las Vegas,
  NV, 2016, pp. 770--778.

\bibitem{deng2009imagenet}
J.~Deng, W.~Dong, R.~Socher, L.-J. Li, K.~Li, and L.~Fei-Fei, ``{ImageNet: A
  large-scale hierarchical image database},'' in \emph{IEEE Conf. Comput. Vis.
  Pattern Recognit.}, 2009, pp. 248--255.

\bibitem{glorot2011deep}
X.~Glorot, A.~Bordes, and Y.~Bengio, ``{Deep Sparse Rectifier Neural
  Networks},'' in \emph{Int. Conf. Artif. Intell. Stat.}, Fort Lauderdale, FL,
  2011, pp. 315--323.

\bibitem{ioffe2015batch}
S.~Ioffe and C.~Szegedy, ``Batch normalisation: Accelerating deep network
  training by reducing internal covariate shift,'' in \emph{Int. Conf. Mach.
  Learn.}, Lille, France, 2015, pp. 448--456.

\bibitem{kingma2015adam}
D.~P. Kingma and J.~Ba, ``{Adam: A Method for Stochastic Optimization},'' in
  \emph{Int. Conf. Learn. Represent.}, San Diego, CA, 2015.

\bibitem{lewis1994heterogenous}
D.~D. Lewis and J.~Catlett, ``{Heterogeneous Uncertainty Sampling for
  Supervised Learning},'' in \emph{Int. Conf. Mach. Learn.}, New Brunswick, NJ,
  1994, pp. 148--156.

\end{thebibliography}

\end{document}